\pdfoutput=1
\documentclass[11pt]{article}

\usepackage[]{ACL2023}

\usepackage{times}
\usepackage{latexsym}
\usepackage{graphicx}
\usepackage{float}
\usepackage{xcolor, soul}
\usepackage{subfig}
\usepackage{booktabs}
\usepackage{pgfplots}
\pgfplotsset{width=8cm,compat=1.9}

\newcommand{\gb}[1]{ { \begingroup\sethlcolor{orange}\hl{GB: #1}\endgroup}}

\usepackage[T1]{fontenc}
\usepackage[utf8]{inputenc}

\usepackage{microtype}

\usepackage{inconsolata}

\usepackage{comment}

\title{Run Like a Girl! \\
Sport-Related Gender Bias in Language and Vision}

\author{Sophia Harrison \\
  Universitat Pompeu Fabra \\
  \texttt{\normalsize sophia.harrisonn@gmail.com} \\\And
  Eleonora Gualdoni \\
  Universitat Pompeu Fabra\\
  \texttt{\normalsize eleonora.gualdoni@upf.edu} \\\And
  Gemma Boleda \\
  Universitat Pompeu Fabra / \\
  \normalsize ICREA\\
  \texttt{\normalsize gemma.boleda@upf.edu} \\}

\begin{document}
\maketitle

\begin{abstract}
Gender bias in Language and Vision datasets and models has the potential to perpetuate harmful stereotypes and discrimination.
We analyze gender bias in two Language and Vision datasets.
Consistent with prior work, we find that both datasets underrepresent women, which promotes their invisibilization.
Moreover, we hypothesize and find that a bias affects human naming choices for people playing sports: speakers produce names indicating the sport (e.g.\ `tennis player' or `surfer') more often when it is a man or a boy participating in the sport than when it is a woman or a girl, with an average of 46\% vs.\ 35\% of sports-related names for each gender.
A computational model trained on these naming data reproduces the bias.
We argue that both the data and the model result in representational harm against women.
\end{abstract}

% to do gemma:
% - go over the paper in general: choose names
% - final version of the abstract (we have to make sure that the statistics don’t change)
% - bullet points for the discussion

\section{Introduction}

%Humans are biased \cite{Haselton2015},

%For the first sentence, please cite: "classics" like caliskan et al and either a couple of surveys if you find them or 4-5 prominent papers (you'll find them in the refs that we already have). 
%For the second sentence, see if you can cite papers that focus more on datasets. If not, no problem, leave it without refs (the ones we cite in the first sentence already cover datasets). In that case, however, add more refs to the first sentence. 
%For the third (bias amplification), see if you can cite other papers apart from the ones we cite already that focus on bias amplification.

Existing social biases and stereotypes against certain groups, such as women and racial minorities, are known to be reproduced by computational models \cite{Caliskan_2017,Bolukbasi2016,hovy-sogaard-2015-tagging,wang-etal-2021-language-models,Blodgett2020}.
This is primarily due to the fact that the datasets that the models are trained on are biased themselves, because, unless explicit steps are taken, datasets tend to mirror social biases \cite{Torralba2011,Rudinger2017,DBLP:conf/naacl/RudingerNLD18}.
%and AI models are ``stochastic parrots" \cite{Bender}. 
Moreover, models often amplify biases, because they over-rely on shallow patterns and lean towards majority labels \cite{Ahmed2022, Deery2022-DEETBD, zhao-etal-2017-men}.

Bias in AI has ethical implications, because it can result in harm for the affected groups: both representational harm, with systems demeaning or ignoring them, and allocational harm, with systems allocating fewer resources or opportunities to them \cite{Barocas2021, Blodgett2020, Mehrabi2022}.
For instance, the fact that the multi-modal model VL-BERT \cite{Su2019} often predicts that a woman carrying a briefcase is carrying a purse \cite{srinivasan-bisk-2022-worst} constitutes representational harm, with working women not being recognized as such.
%\cite{kurita-etal-2019-measuring} and viceversa \cite{https:Leino2018}.
% srinivasan and bisk:
%Relying on stereotypical cues
%(learned from biased pre-training data) can cause
%the model to override visual and linguistic evidence
%when making predictions. 

% srinivasan and bisk:
%Several papers have
%investigated how dataset biases can override visual
%evidence in model decisions. \cite{zhao-etal-2017-men}
%showed that multimodal models can amplify gender
%biases in training data. In VQA, models make
%decisions by exploiting language priors rather than
%utilizing the visual context \cite{Goyal2017, Ramakrishnan2018}. Visual biases can also
%affect language, where gendered artifacts in the
%visual context influence generated captions \cite{Hendricks2018, Bhargava2019}

% humans can show a bias in their lexical choice \cite{Silberer2020, Torralba2011, Wen2021}.  

\begin{comment}

\begin{figure}[htb]
  \centering
  \subfloat[
  \textbf{woman (16)}, surfer (14)]{\includegraphics[width=0.47\columnwidth]{Images/surfer_woman.png}\label{fig:surfer_woman}}
    \hfill
  \subfloat[
  \textbf{surfer} (24), man (6), person (2), boy (2)]{\includegraphics[width=0.47\columnwidth]{Images/surfer_man.png}\label{fig:surfer_man}}
\label{fig:surfers}
\end{figure}

\end{comment}

\begin{figure}[htb]
  \centering
  \subfloat[
  \textbf{woman} (16), surfer (14)]{\includegraphics[width=0.47\columnwidth]{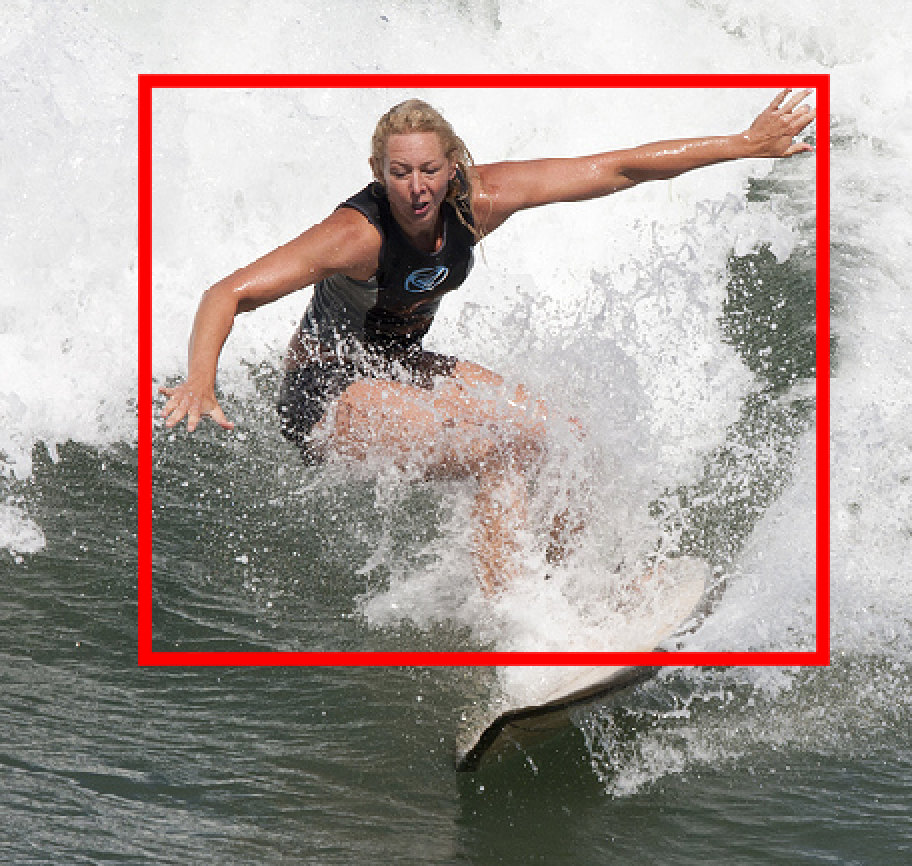}\label{fig:surfer_woman}}
    \hfill
  \subfloat[
  \textbf{surfer} (24), man (6), person (2), boy (2)]{\includegraphics[width=0.47\columnwidth]{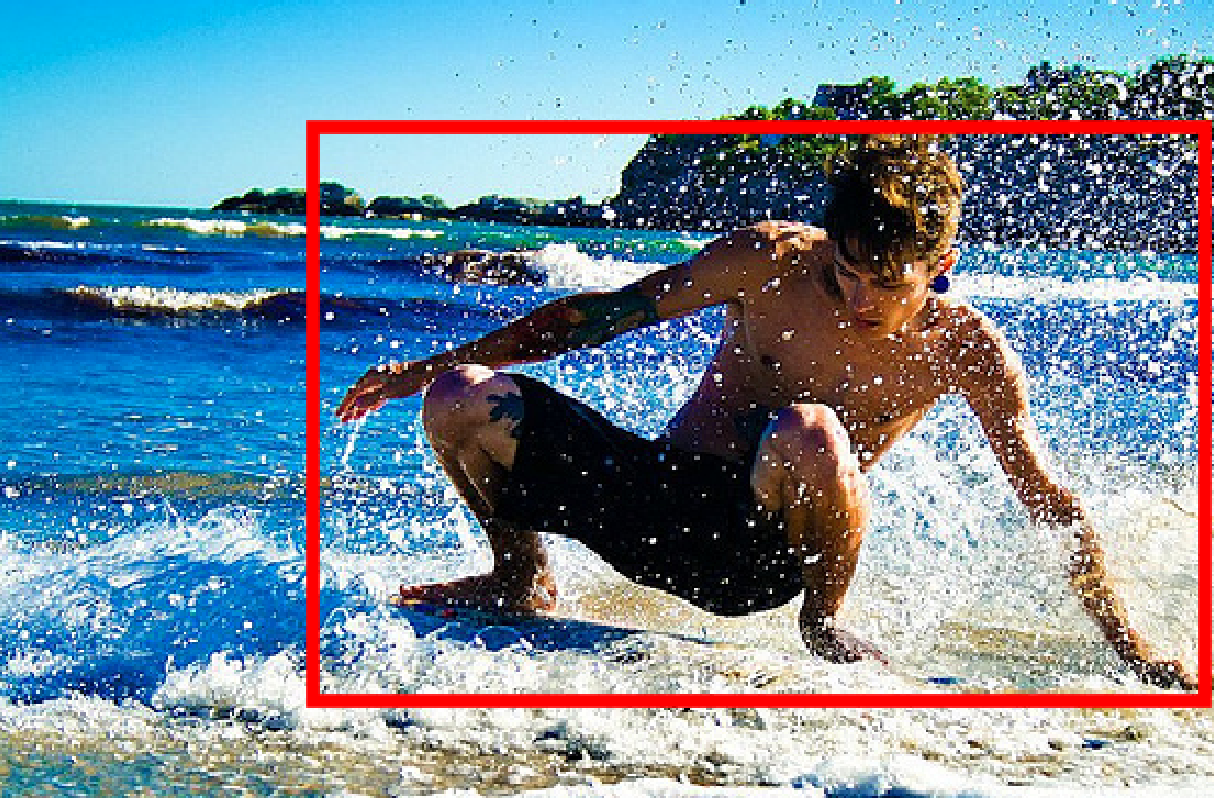}\label{fig:surfer_man}}
  \hfill
  \subfloat[
  \textbf{girl} (11), skateboarder (7), skater (6), child (6)]{\includegraphics[width=0.47\columnwidth]{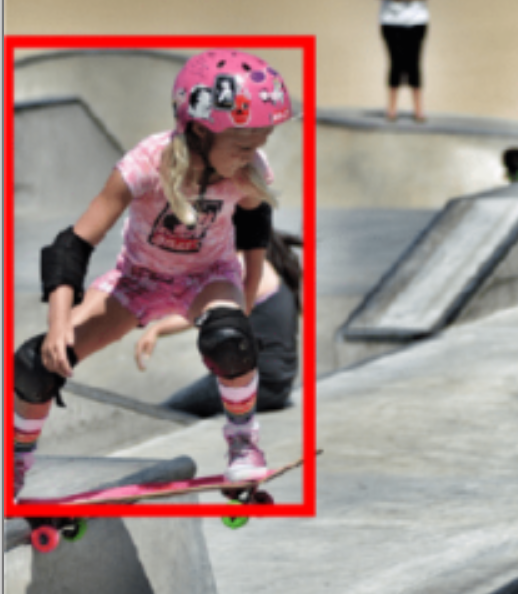}\label{fig:skater_girl}}
   \hfill
  \subfloat[
  \textbf{skater} (9), skateboarder (6), boy (6), kid (3)]{\includegraphics[width=0.47\columnwidth]{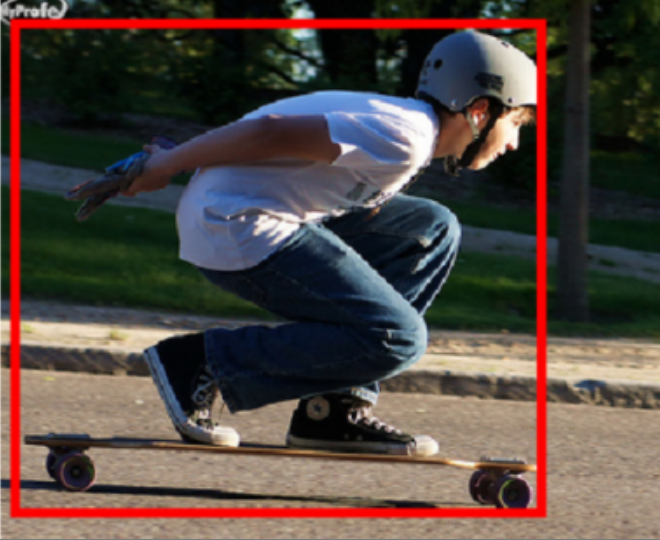}\label{fig:skater_boy}}
  \caption{Images of people playing sports from the ManyNames dataset, together with the names that human annotators produced and their counts.}
   \label{fig:surfers}
\end{figure}

In this paper, we focus on gender bias that causes representational harm for women, and specifically for women in sports, in the area of Language and Vision. 
Previous work on gender bias in L\&V has shown that bias often relates to the language component, causing models to override the specific visual information in classification decisions \cite{zhao-etal-2017-men,Goyal2017, Ramakrishnan2018} and vice versa \cite{Hendricks2018, Bhargava2019}, such as in the `purse/briefcase' example above. We examine bias that is present in the language, without examining its interaction with the visual component in the model. 

Women in Western societies have traditionally been marginalized, excluded, and deterred from participating in sports or even physical activity, while men have been encouraged \cite{Bell2008,Scheadler2018, Vertinsky1994, Schaillée2021}. 
Sport has long been stereotypically associated with masculinity, and females have been thought physically incapable of performing well in this area \cite{Young1980}. 
Participation in sports by women and girls has continued to increase since the 1960s, in parallel with other social advances; however, rarely do women rise to managerial or coaching roles, even for women's teams \cite{Schaillée2021}.
Women's sports are also underrepresented in the media, which has been linked to perpetuating the stereotype of the male athlete over the female \cite{Schmidt2013}.

We analyze the data in a Language and Vision dataset for object naming, ManyNames (\citealp{Silberer2020,silberer2020_MN2}; more information in the next section). Figure \ref{fig:surfers} shows two example images in ManyNames together with the names elicited from subjects. How we name an object or entity is intimately linked to how we conceptualize it \cite{Brown1958HowSA,LaTourrette2020}.
We hypothesize that, due to the social constraints discussed above, speakers produce a sports-related name such as `surfer' less often when their referent is a woman, that is, they do not conceptualize female athletes as athletes. 
If the bias indeed exists in the speaker population, we expect it to be present in the naming data and propagate to computational models; we check both expectations. 
Before that, we examine whether there is an overall representational bias in ManyNames and its parent dataset, VisualGenome \cite{Krishna2016}, with women being underrepresented. 

\section{Underrepresentation of Women}
\label{sec:underrepresentation}

\paragraph{Data and method} 

ManyNames contains names for 25K images produced by English-speaking subjects in a free naming task, with an average of 31 names per image. 
ManyNames images are a subset of those in VisualGenome. The images were selected from seven previously defined domains, one of which was PEOPLE, based on a series of seed WordNet synsets. The authors put a cap on the number of images for a given synset (max.\ 500 instances for seeds with up to 800 objects in VisualGenome, and up to 1k instances for seeds with more than 800 objects). 

VisualGenome contains 108K images that are the intersection of images in the datasets YFCC100m \cite{Thomee2016} and MS-COCO \cite{lin2015}. 
The objects in each VisualGenome image were manually identified, labeled, and linked to their corresponding WordNet synset (VisualGenome provides many more annotations, but these are the ones of interest for the present study). 
Both VisualGenome and ManyNames employed crowd-source workers from Amazon Mechanical Turk (AMT) as annotators \footnote{In VisualGenome, annotators identified (via bounding boxes) the objects in each image, and provided descriptions such as "a red mushroom with white spots". The head nouns in these region descriptions were identified and matched with WordNet synsets in a semi-automatic fashion. In ManyNames, annotators were shown images where an object was highlighted with a bounding box, and they were asked to provide a single name for the object. 36 different subjects provided names for each object, and the naming responses were cleaned and aggregated. We use the object synsets for the analysis in Section 2, and the naming distributions for the rest of the paper.} for the images, with workers coming predominantly from the USA. Note that while the images in ManyNames are a subset of those in VisualGenome, the naming annotations were collected afresh for ManyNames.

%For ManyNames, the workers were from Western countries (USA, Canada, UK, New Zealand, and Australia); for VisualGenome, 93.02\% were from the USA and the remaining 7\% were from other countries (Philippines, Kenya, India, Russia, Canada, and others).

The YFCC100m images were all those uploaded to Flickr between 2004 and 2014, published under a commercial or non-commercial license. MS-COCO contains all images from Flickr available at the time of the dataset construction that belonged to 91 predefined image categories (e.g.,\ \textit{horse}, \textit{people}, and \textit{laptop}). Images on Flickr are uploaded by Flickr users. 
%\footnote{The annotation process involved multiple rounds where workers described image regions and drew bounding boxes around these areas. After the image regions were drawn and described, annotators labeled objects based on the region description given to the image by a human annotator. They also drew bounding boxes around the specific objects mentioned in the region.} with a single hard-truth label name, along with captions and region descriptions. 

To check for representational bias, we extracted all objects in VisualGenome (image areas within a bounding box) with labels corresponding to the four most common gender-associated names: `boy', `girl', `man', and `woman'. For ManyNames, we used the same names and the full naming distribution. 

\begin{table}[htb]
\centering
\begin{tabular}{cccc}
 \textbf{Names} & \textbf{VG} &   \textbf{MN} & \textbf{World}\\ \toprule
woman & 25.6 &  39.7   & -\\
girl & 7.0 &  7.9  & -\\
man & 59.1 & 37.9  & - \\
boy & 8.1 & 14.4   & - \\ \hline
total female & \textbf{32.7} & \textbf{47.6}  & \textbf{49.6} \\
total male & 67.2 &  52.3 & 50.4\\
%\verb|#| images & 147,875 &  3,630  & -\\ 
\bottomrule
\end{tabular}
\caption{Gender distribution in VisualGenome (VG), ManyNames (MN), and the world in 2020, in percentage.}
\label{tab:study1}
\end{table}

\paragraph{Results} 
The resulting gender distribution is shown in Table~\ref{tab:study1}, together with the distribution in the world in 2020 as reported by the United Nations world population data \cite{UnitedNations2020}.
Both datasets indeed underrepresent females, according to one-tailed z-tests comparing the percentage of females in each dataset with the global percentage of females in the world (VisualGenome: z = -129.8, p < 0.001; ManyNames: z = -2.4, p = 0.02).%
\footnote{All statistical tests in this paper assume an alpha level of 0.05.}
Note, however, that the bias against women is much larger in VisualGenome, with only 32.7\% female entities compared with 49.6\% in the world's population; in ManyNames, the percentage is only 2 points below the world population (47.6\% vs 49.6\%). 
Moreover, note that the bias in ManyNames stems from images of boys, with 14.4\% of images vs 7.9\% for girls. 
Recall from above that no specific action was taken in either ManyNames or VisualGenome regarding gender balance. The representational bias in ManyNames is smaller due to the cap on the synsets, aimed at obtaining a varied set of categories in general; the reduced gender bias is a side effect.
Below we discuss a further specific type of underrepresentation that we also find in ManyNames, that of women playing sports. 

\section{Sport-Related Bias}
\label{sec:naming_bias}

We next present our main analysis, namely gender bias related to sport as shown in human naming data. A secondary analysis concerns model behavior. 

\paragraph{Sport-related bias in humans: Methods}

The first author of the paper went through all topnames---that is, the name produced by the majority of the annotators for each image---in the ManyNames domain PEOPLE and selected those that related directly to gender (`boy', `girl', `man', `woman') or sport (`athlete', `baseball player', `basketball player', `batter', `catcher', `goalie', `pitcher', `player', `skateboarder', `skater', `skier', `soccer player', `snowboarder', `surfer', `tennis player', `umpire'). 
We refer to the former as "taxonomic" and the latter as "sport-related".
We selected all  1,776 images that have at least one taxonomic and one sport-related name in the responses, such that we could automatically determine both the gender of the person and the fact that they are playing a sport.%
\footnote{Images with inconsistent gender in the response (e.g., where some subjects produced `man' and others `woman') were discarded.}

To check for bias, we computed, for each image, the percentage of sport-related names associated with it, relative to the total names, which include both taxonomic and sport-related names. 
For instance, in Figure \ref{fig:surfers}, the person in panel (a) received 46.7\% of sport-related names, while the person in panel (b) received 70.5\%.
We fitted a logistic regression model with the proportion of sport-related names as the outcome variable and fixed effects for the person's gender. 

\begin{figure}[H]
  \centering
  \includegraphics[width=.8\columnwidth]{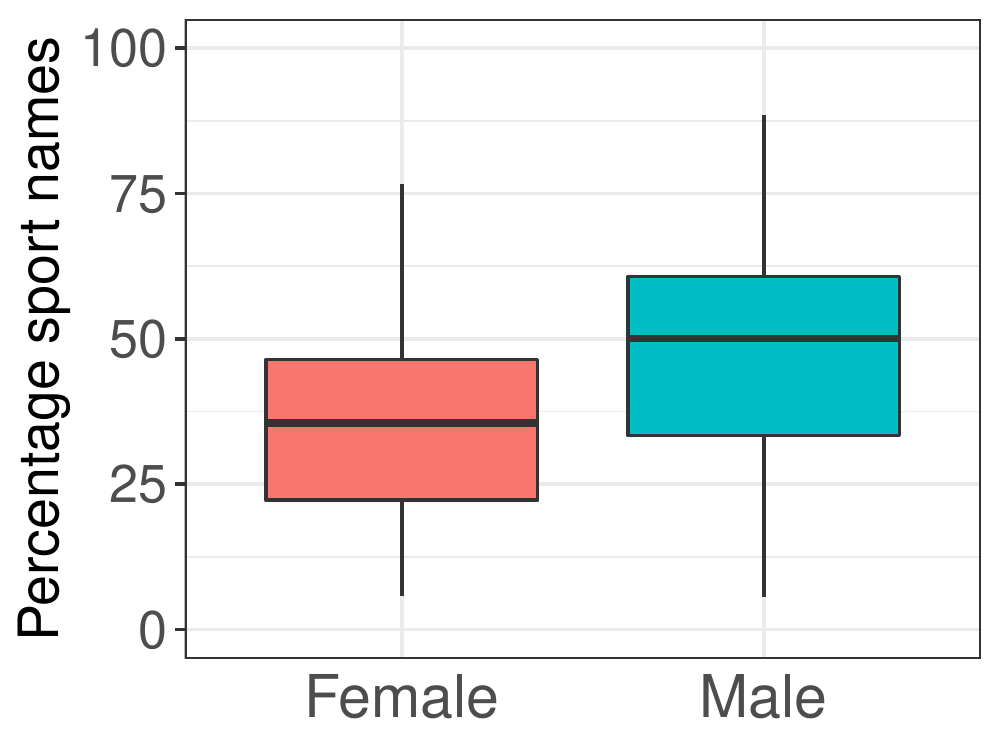}
  \caption{Distributions of percentages of sport-related names for female and male images in ManyNames.}
    \label{fig:updated_plot}
\end{figure}

\paragraph{Sport-related bias in humans: Results}
Table \ref{tab:regression_human_bias} summarizes the results of the logistic regression, supporting our hypothesis: when annotators see images of men playing sports, they are more likely to produce a name that explicitly mentions the sport being played, and therefore are less likely to produce a taxonomic name, compared with when they see images of women playing sports.
Figure \ref{fig:updated_plot} qualitatively shows the difference between the genders. 
Note that there are also fewer images of women playing sports (527 vs.\ 1219), constituting only the 30.2\% of the pictures of people playing sports.
These numbers compared to the real-world sports statistics constitute a further, more insidious instance of underrepresentation of women in L\&V datasets: women in certain roles. This constitutes representational bias regarding women playing sports in ManyNames. Note that there are in general fewer women playing most sports in the Western population---especially due to dropouts \cite{Bevan2021}; however, the difference between the genders is not as large as in the dataset. For instance, in the US, the percentage of girls in college sports in 2019 was 43.9\% \textit{vs} 56.1\% boys \cite{NCAA2022} and in England the percentage of women among the adults who participated in a sports activity in 2020 was 45\% \textit{vs} 55\% men \cite{SportEngland}. 
%Of note, the gender gap in sports has reduced greatly in the last few decades: in 1972, only 7\% of high school athletes were female and in 1982 this percentage became 27.8\% (NCAA, 2022). A tentative hypothesis can be that social biases reflected in language lag behind social change.

%---such that this does not necessarily correspond to underrepresentation.

\begin{table}[!htbp] \centering 
\begin{tabular}{lccc}
    \toprule 
 \multicolumn{4}{c}{Model; dep.\ var.: Prop sport-related names} \\ \hline
   & Estimate & St. Error & p-value\\
 Intercept & $-$0.62 & 0.02 & p$<$0.001 \\ 
 GenderM & 0.47  & 0.02 & p$<$0.001\\  
 \midrule 
 \multicolumn{4}{c}{Descriptive statistics: Perc sport-related names} \\ \hline
   & N & M & SD\\ 
 female & 527 & 34.9 & 16.9\\ 
 male & 1219 & 46.2 & 18.7\\  
 \bottomrule
\end{tabular} 
\caption{Top: Fixed-effect estimates for the logistic regression model predicting the proportion of sport-related names based on image gender. Bottom: descriptive statistics of the sample (N = number of images; M = mean; SD = standard deviation).} 
\label{tab:regression_human_bias}
\end{table} 

\paragraph{Sport-related bias in an L\&V model}

We additionally analyze the behavior of a model trained to produce names on the ManyNames dataset \cite{Lang2021}. Based on the extensive literature on bias \cite{mehrabi-etal-2022-robust}, we expect the model to reproduce the biases observed in the data; in contrast, it is unclear whether it will amplify it, as models vary with respect to this \cite{zhao-etal-2017-men, fernando2021missing}. 

This model builds upon a ResNet101 architecture \cite{He2015} pretrained on VisualGenome \cite{Anderson2018} and is adapted to ManyNames names through an additional fine-tuning step. 
Relevantly to our purposes, the model is trained to reproduce the full naming distribution, outputting a probability distribution over all the names in the vocabulary. 
This is different from object classification in Computer Vision, which assigns a single class to each object \cite{Deng2009, faster, He2015}.
%
%\footnote{The models were trained to focus on three features of an image: the entirety of the image itself, taking context, object, and the whole scene into account; the target object only, and the context only \cite{Lang2021}.}} 
%The goal of our analysis is to determine whether the sport-related gender bias found in the ManyNames naming data is reproduced, or possibly even amplified, by the model. 
\citet{Lang2021} used the train--dev--test partition of ManyNames established in \citet{silberer2020_MN2}; we analyze the behavior of the model in the test set and, in particular, in the images that meet the criteria used in our second analysis above (N = 89, of which 34 female).

We normalize the probability weights output by the model, using only the names of interest (that is, discarding any weight the model places on names that are neither taxonomic nor sport-related before doing the normalization). 
To check whether the model reproduces the bias, we fit a logistic regression model, with the proportion of sport-related names as the dependent variable and fixed effects for the image gender. 
To check whether it amplifies the bias, we fit a mixed-effects logistic regression model that takes into account both the model and the human data (see Appendix for details). 
\footnote{Models were fit in R using {\em glm} and {\em glmer} \citep{glmer, R}.}

As shown in Table \ref{tab:regression_model_bias}, according to the regression analysis, the L\&V model indeed reproduces the naming bias found in the human data: for images depicting boys and men playing sports, it assigns significantly higher weights to sport-related names than for images depicting women or girls playing sports. 
We instead find no evidence of bias amplification (see Appendix); note however that the sample is small.

%a t-test comparing model and human naming data for females yielded no significant results (p = 0.09), with model M = 36.6 and SD = 21.9, and humans M = 28.2 and SD = 18.4. 

%\begin{table}[]
%\centering
%\begin{tabular}{@{}ccc@{}}
% \textbf{Model} & Taxonomic & Sport-related \\ %\toprule
 %Male & \EG{ADD} & $48.5\%$ ($\pm 24.5\%$) \\
 %Female &  \EG{ADD} & $36.3\%$ ($\pm 21.9\%$) \\ %\midrule\midrule
 %\textbf{Humans} & Taxonomic & Sport-related \\ %\toprule
% Male & \EG{ADD} & \EG{ADD} \\
% Female &  \EG{ADD} & \EG{ADD} \\
% \bottomrule
%\end{tabular}
%\caption{Average percentages of taxonomic and sport-related names for the images produced by model and humans for the images included in our third study, divided by gender.}
%A t-test shows that there is a statistically significant difference (p = 0.02) between the distribution of probability weights given by the model to images of males playing sports (M = 48.5, SD = 24.5) and the distribution given to images of females playing sports (M = 36.6, SD = 21.9). 
%\label{tab:bias_model}
%\end{table}

\begin{table}[!htbp] \centering 
\begin{tabular}{lccc}
    \toprule 
 \multicolumn{4}{c}{Model; dep.\ var.: Prop sport-related names} \\ \hline
   & Estimate & St. Error & p-value\\
 Intercept & $-$0.62 & 0.07 & p$<$0.001 \\ 
 GenderM & 0.63  & 0.08 & p$<$0.001\\  
 \midrule 
 \multicolumn{4}{c}{Descriptive statistics: Perc sport-related names} \\ \hline
   & N & M & SD\\ 
 female & 34 & 36.6 & 21.9\\ 
 male & 55 & 48.5 & 24.5 \\  
 \bottomrule
\end{tabular} 
\caption{Top: Fixed-effect estimates for the logistic regression model predicting the proportion of sport-related names based on image gender. Bottom: descriptive statistics of the sample (N = number of images; M = mean; SD = standard deviation).} 
\label{tab:regression_model_bias}
\end{table}

\section{Discussion and conclusions}

We have identified pervasive biases against women in Language \& Vision.
Our main contribution is the individuation of a bias that characterizes human naming choices and therefore the naming data available for models. While we have focused on ManyNames, this issue is likely to affect other datasets containing names, such as widely used datasets for captioning \cite{flickr, concap, lin2015} and referring expression generation \cite{referit, refcoco}. 
The bias concerns the kind of name chosen for athletes, depending on the genre: people produce fewer sport-related names for females playing sports (average 35\%) than for males playing sports (46\%).
As far as we know, this kind of bias has not been previously discussed, and it is, we argue, more implicit and thus difficult to identify than other kinds of bias that are more commonly discussed in the literature, such as unbalanced classes in datasets due to the underrepresentation of certain groups (which we have also found; discussion below). 
We find the naming bias both in the human data of ManyNames and a model trained on this data. 
Thus, even women that do play sports (which are fewer to begin with, as mentioned in Section~\ref{sec:naming_bias}) are not conceptualized as such. 
This constitutes representational harm and contributes to limiting choices for women. 
% Naming images of males by a sport-name more often than images of females may reflect an implicit bias of our society, intimating that playing a sport is a masculine behavior. This bias has deep roots: Sports are historically and culturally segregated \cite{Dennis2015} in addition to being one of the most observable manifestations of gender bias and discrimination.

We also find that women are underrepresented in L\&V, especially pronounced in VisualGenome, which has over twice as many images of males than females. 
Moreover, the proportion of males and females playing sports in ManyNames is skewed, with only 30.2\% of the pictures of people playing sports depicting females. 
As mentioned above, the actual percentage of women and girls in sports in Anglosaxon societies is closer to 45\%, according to recent data \cite{NCAA2022,SportEngland}.
Underrepresentation of social groups is harmful in itself \cite{Blodgett2020}, and also because models trained on unbalanced data can neglect crucial patterns relevant to those within the group \cite{Wang_2022}. %Concomittantly, underexposure of minority data to models can lead to discrimination \cite{huvy
%VisualGenome, as well as the datasets whose images come from, YFCC100m and MS-COCO, are widely used. 
% While the authors of VisualGenome mention that``Due to the image biases that exist in the dataset, we have twice as many annotations for men (24k) than we do of women(11k)'' \cite{Krishna2016}, they do not discuss the origin of the biases. 
Given how the images were selected (see Section~\ref{sec:underrepresentation}), the origin of the underrepresentation of women in these datasets must come from the kinds of images uploaded onto Flickr around the 2010s. 
This is in turn likely rooted in the demographic characteristics of Flickr users in that period, and to the fact that, in general, the internet has been heavily male-dominated \cite{Mohahan-Martin1998, flickr-male}. 
This is a serious concern, as most resources used in L\&V (as well as computational linguistics and AI in general) come from the internet. 
%As mentioned above, the bias is mitigated in ManyNames due to the cap imposed by the authors on the number of images per label (which was aimed at achieving variety in ManyNames, not at avoiding bias per se, \citealp{Silberer2020}). %However, the persistence of the bias can be explained through the interplay of taxonomic and sport-related names. Images pulled from VisualGenome that were labeled with a sport-related name, such as 'skater', did not highlight the gender of the person in the image, making it hard for the authors of the dataset to control for the image gender. %In our study, thanks to the multiple we labeled all images of people playing sports with a gender based on the names given by human-annotators. 
% Interestingly, however, the bias is a result of the images of 'boy' and 'girl'. There are slightly more images of boys in ManyNames than of girls, ultimately creating a skew towards images of males. Furthermore, ManyNames pulled images labeled by sport-role. These sports images were not initially labeled by a taxonomic name, however, during object naming, human annotators identified many of these images of people playing sports by their taxonomic name. Then for our study we separated all images of people playing sports into male or female, thus discovering the actual image count of males to be higher than initially accounted for. 

Our findings are in line with other research on gender bias, both in L\&V and in NLP and AI more broadly, discussed in the introduction. They also resonate with a study of the Pew Research Center \cite{Pew} showing that results in Google Image Search underrepresent women in various jobs, compared to their actual participation in those jobs in the USA according to the Bureau of Labor Statistics. 

Ultimately, based on the findings of this paper, it can be concluded that, as far as the datasets and model analyzed are concerned, when it comes to sports: A man with a tennis racket is a tennis player. A woman with a tennis racket is just a woman with a tennis racket.

\section*{Limitations and Future Work}

Our findings about gender bias in the field of Language \& Vision are based on two datasets, one task (Object Naming), one language (English), a mostly Western population (based on the origin of both the images and the annotators of VisualGenome and ManyNames), and one computational model. 
Also, in the third analysis, due to the characteristics of the test set of ManyNames, the sample size was small. 
Moreover, the bias around naming choices concerns the domain of sports only. 

Regarding our most novel finding (bias in lexical choice), given the basic function of naming in language, and the fact that Western English-speaking societies are not known to be more gender-biased than most non-Western and/or non-English speaking societies, it is plausible that the identified bias extends to other L\&V tasks such as image captioning, referring expression generation, or Visual Question Answering.
Also given previous work on bias in our field, it is plausible that the identified bias in the model extends to other models. It is however not clear whether the bias will be amplified or simply reproduced. 
To probe whether, and to what extent, the identified biases indeed generalize, future work should tackle more tasks, languages, populations, domains, models, and data. 
Testing further models on the same naming data that we have used is straightforward; checking for biases in some other tasks for English should be feasible at least to some extent, since some datasets provide multiple annotations per image (e.g.\ captions in MS-COCO). Instead, analyzing other languages and populations, such as non-binary individuals, will in most cases require further data collection, due to the scarcity of non-WEIRD data in our field. 

In this study, gender was operationalized in a binary manner. There is a lack of gender-neutral labels within the datasets used (5\% of labels can be considered gender-neutral, i.e., "person, human, child"), and the resources required to reflect the reality of the gender landscape currently do not exist. This is a symptom of a further issue related to gender bias, namely a lack of representation of non-binary individuals within vision datasets and the difficulties of conducting ethically inclusive studies on gender \cite{larson2017gender}. Addressing this is beyond the scope of the present research and remains an important direction for future work. Finally, this work solely concerns the identification of biases; further work should focus on how to deal with them  in terms of data collection, curation, and modeling.

\section*{Ethics Statement}
This research aims to highlight the extent to which gender biases may be present in Language \& Vision, with an emphasis on the representation of females in sports. The findings of the study may have implications for the ways in which images of female athletes and sports figures are portrayed and treated within the datasets and by models. The results of the study may have the potential to contribute to the ongoing efforts to address gender bias in all areas of life, as we maintain that gender bias is an issue within Machine Learning because it is an issue within human society. The research team is committed to using the findings of the study to foster dialogue and understanding of the issue, and to advocate for equitable treatment of all groups throughout the production pipeline.

\section*{Acknowledgments}

The authors thank the reviewers for their useful feedback. This research is an output of grant PID2020-112602GB-I00/MICIN/AEI/10.13039/501100011033, funded by the Ministerio de Ciencia e Innovación and the Agencia Estatal de Investigación (Spain) and of grant agreement No.\ 715154 funded by the European Research Council (ERC) under the European Union's Horizon 2020 research and innovation programme. This paper reflects the authors' view only, and the funding agencies are not responsible for any use that may be made of the information it contains.

\begin{flushright}
\includegraphics[width=0.8cm]{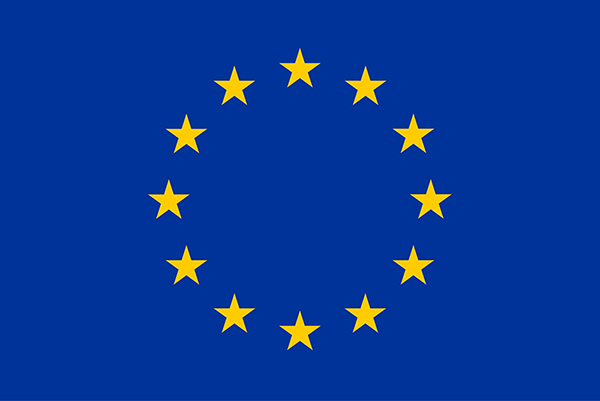}
\includegraphics[width=0.8cm]{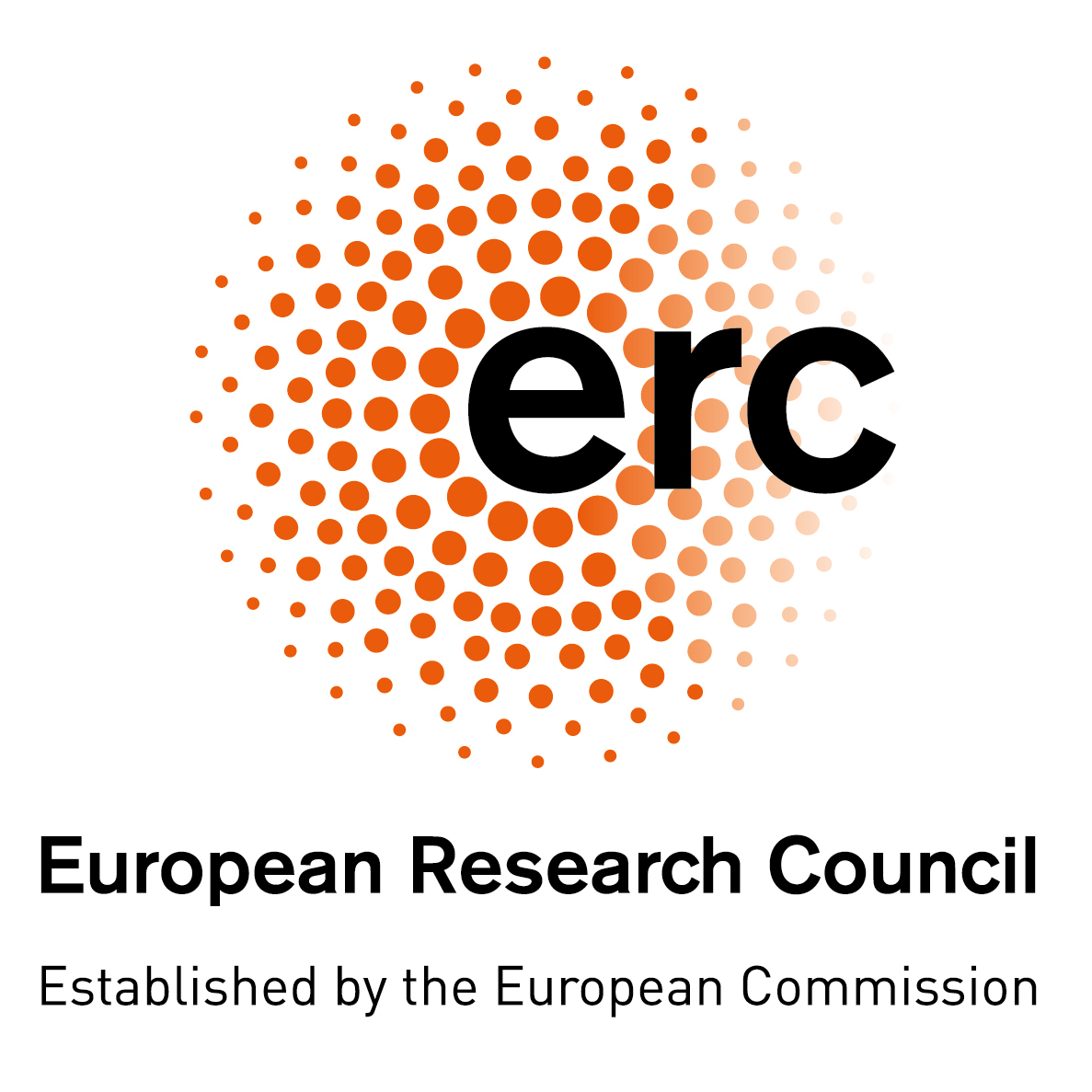}
\end{flushright}

\bibliography{anthology}
\bibliographystyle{acl_natbib}

\appendix

%\section{Boxplot blabla}
%\label{sec:appendix}

%\begin{figure}[H]
%  \centering
%  \includegraphics[width=\columnwidth]{updated_model_plot.png}
%  \caption{ Distribution of model probability weights assigned to sport-related names, for images of
%males and females playing sports, compared with the distribution of model probability weights assigned to taxonomic names for the same images.
%plot 1: Model Probability for Female Images
%plot 2: Model Probability for Male Images
%plot 3: Human Percentages for Female Images
%plot 4: Human Percentages for Male Images}
%    \label{fig:updated_model_plot.png}
%\end{figure}

\begin{comment}

\section{Further illustration of the bias}

\gb{ONLY AFTER YOU'VE DONE EVERYTHING ELSE: @Sophia pls add some text and finish the figure. You'll need to change image names,  to play with the widths of the images.}

\begin{figure}[htb]
  \centering
  \subfloat[
  \textbf{XXX}]{\includegraphics[width=0.47\columnwidth]{Images/surfer_woman.png}}
    \hfill
  \subfloat[
  \textbf{YYY}]{\includegraphics[width=0.47\columnwidth]{Images/surfer_man.png}}
  \\
  \subfloat[
  \textbf{ZZZ}]{\includegraphics[width=0.3\columnwidth]{Images/girl_skateboarding}}
    \hfill
  \subfloat[
  \textbf{WWW}{\includegraphics[width=0.4\columnwidth]{Images/boy_skateboarding.png}}
  \caption{Images of people playing sports from the ManyNames dataset, together with the names and that human annotators produced and their counts.}
    \label{fig:surfers}
\end{figure}

\end{comment}

\section{Bias amplification}
\label{app:model_amplification}

To check whether the model amplifies the bias, we considered, for the same 89 images, the probabilities assigned to taxonomic and sport-related names by the human annotators and the model and
fitted a mixed-effects logistic regression model, with the proportion of sport-related names as the outcome variable and fixed effects for the image gender, the prediction type (either \textit{human} or \textit{model}), and the interaction between image gender and prediction type. We set a random intercept for each image and a random slope for each type. We coded as ``treatment" the prediction type so that the \textit{human} prediction would be our baseline.

%Here we report the results of the mixed-effects logistic regression model fitted to test whether the L\%V model amplifies the gender bias. 

\begin{table}[!htbp] \centering 
\begin{tabular}{cccc} \\
 \multicolumn{4}{c}{\textit{Dependent variable:} Prop Sport-related Names} \\ 
\toprule 
   & Estimate & St. Error & p-value\\ 
 Intercept & -1.05 & 0.15 & p$<$0.001 \\ 
 GenderM & 0.99 & 0.19 & p$<$0.01\\  
 TypeM & 0.25 & 0.20 & p$<$1 \\
GeM:TyM & -0.26 & 0.263 & p$<$1 \\
 \midrule 
 \multicolumn{4}{c}{Descr.\ stats., model: Perc sport-related names} \\ \hline
   & N & M & SD\\ 
 female & 34 & 36.6 & 21.9\\ 
 male & 55 & 48.5 & 24.5 \\  
 \midrule 
 \multicolumn{4}{c}{Descr.\ stats., human: Perc sport-related names} \\ \hline
   & N & M & SD\\ 
 female & 34 & 28.2 & 18.3\\ 
 male & 55 & 49 & 18.3 \\  
 \bottomrule
\end{tabular} 
\caption{Fixed-effect estimates for mixed-effects logistic regression model predicting the proportion of sport-related names based on image gender and type of prediction (human \textit{vs} model - with human treated as baseline).} 
\label{tab:regression_model_ampli}
\end{table}

The results are summarized in Table~\ref{tab:regression_model_ampli}. The regression shows that, for the 89 images included in this analysis, humans use more sport-related names when the image gender is male --in line with our previous findings. However, the results concerning a possible bias amplification are inconclusive: The estimates for the prediction type and for the interaction between type and gender are not significant. Analyses of larger data samples may be needed to shed further light on this topic.

\end{document}